%% file: main.tex
\documentclass[10pt, a4paper]{article}
\usepackage{lrec}
\usepackage{multibib}
\newcites{languageresource}{Language Resources}
\usepackage{graphicx}
\usepackage{tabularx}
\usepackage{soul}
\usepackage{subfiles}
\usepackage{epstopdf}
\usepackage[utf8]{inputenc}
\usepackage{CJKutf8}
\usepackage{booktabs,caption}
\usepackage[flushleft]{threeparttable}

\usepackage{hyperref}
\usepackage{xstring}
\usepackage{url}
\usepackage{color}
\usepackage{multirow}
\usepackage[T2A,T1]{fontenc}
\usepackage{changepage}
\usepackage{enumitem}

\renewcommand\thesection{\arabic{section}}
\renewcommand\thesubsection{\thesection.\arabic{subsection}}

\title{Using Linguistic Typology to Enrich Multilingual Lexicons:\\the Case of Lexical Gaps in Kinship}

\name{Temuulen Khishigsuren$^{1}$, Gábor Bella$^2$, Khuyagbaatar Batsuren$^{1}$, \\ {\bf \large Abed Alhakim Freihat$^2$, Nandu Chandran Nair$^2$, Amarsanaa Ganbold$^1$,} \\ {\bf \large Hadi Khalilia$^2$, Yamini Chandrashekar$^2$, Fausto Giunchiglia$^2$}}

\address{$^{1}$National University of Mongolia, Mongolia, $^2$University of Trento, Italy \\
         kh.temulen@gmail.com, gabor.bella@unitn.it,
         khuyagbaatar@num.edu.mn,\\
         abdel.fraihat@gmail.com,
         nandu.chandrannair@unitn.it,
         amarsanaag@num.edu.mn,\\
         hadi.khalilia@unitn.it,
         yamini.chandrashekar@unitn.it,
         fausto.giunchiglia@unitn.it
         \\}

\abstract{
This paper describes a method to enrich lexical resources with content relating to linguistic diversity, based on knowledge from the field of lexical typology. We capture the phenomenon of diversity through the notions of \textit{lexical gap} and \emph{language-specific word} and use a systematic method to infer gaps semi-automatically on a large scale. As a first result obtained for the domain of kinship terminology, known to be very diverse throughout the world, we publish a lexico-semantic resource consisting of 198~domain concepts, 1,911~words, and 37,370~gaps covering 699~languages. We see  potential in the use of resources such as ours for the improvement of a variety of cross-lingual NLP tasks, which we demonstrate through a downstream application for the evaluation of machine translation systems. \\ \newline  }

\begin{document}

  \maketitleabstract

\section{Introduction}
\subfile{sections/1.intro.tex}

\section{Untranslatability and Lexical Typology}
\label{sec:our_proposal}
\subfile{sections/3.lexgap.tex}

\section{Lexical Gap Generation Method}

\label{sec:method}
\subfile{sections/4.method.tex}

\section{Evaluation}
\label{sec:evaluation}
\subfile{sections/5.evaluation.tex}

\section{The Published Resource}
\label{sec:result}
\subfile{sections/6.result.tex}

\section{Application in Machine Translation}
\label{sec:application}
\subfile{sections/7.application.tex}

\section{Related Work}
\label{sec:soa}

\subfile{sections/2.related.tex}

\section{Conclusion and Future Work}
\label{sec:conclusion}
\subfile{sections/8.conclusion.tex}

\section{References}
\bibliographystyle{lrec}
\bibliography{main}

\section{Language Resource References}
\label{lr:ref}
\bibliographystylelanguageresource{lrec}
\bibliographylanguageresource{languageresource}

\end{document}

%% file: sections/1.intro.tex
To address the language technology bottleneck beyond a handful of well-supported languages, the computational linguistics community has proposed several solutions: unsupervised learning models that remove the need for parallel textual corpora \cite{snyder2008unsupervised,artetxe2017unsupervised,radford2019language} or cross-lingual transfer from high- to low-resourced languages \cite{hwa2005bootstrapping,pado2005cross,tackstrom2012cross}. Joint supervised learning and representation learning also happen to be effective in certain multilingual NLP applications, e.g.~neural machine translation \cite{ammar2016many,guo2016representation}.

The common point of these methods is that they rely on an implicit assumption of \emph{sameness} across languages: for example, that in cross-lingual transfer the lexicon of one language can be efficiently mapped to that of another language through a shared semantic vector space. While the existence of shared cross-lingual conceptualizations is obvious to all---otherwise no interlingual communication could ever be possible---the world's languages are, however, also known to be extremely \emph{diverse} on every level \cite{levinson2010time}. Thus, in recent years there has been an emerging trend to exploit results from the field of \emph{linguistic typology} into language diversity in various multilingual NLP tasks (for a review, please see \newcite{ponti2019modeling,arora2022computational}). Typology-informed NLP studies use typological features from hand-curated resources, such as the World Atlas of Language Structures (WALS) \citelanguageresource{wals}, to bind parameters of languages that have similar typological features within the framework of cross-lingual transfer \cite{tackstrom2012cross} or to manipulate joint supervised learning models to reflect typological features of specific languages \cite{ammar2016many,bjerva2020sigtyp}. The use of cross-lingual cognate databases for bilingual lexicon induction was explored by \newcite{batsuren2021large}. The use of typology in various NLP applications generated consistent improvement, and is thus regarded as an effective way to advance the development of multilingual NLP \cite{cotterell2019complexity,salesky2020corpus,pimentel2021surprisal}. 


Most typology-informed NLP studies, however, are limited to recognizing language-specific morphosyntactic features and have so far ignored diversity within lexicons. Yet, the vocabularies of languages differ considerably in their division of the semantic space. For example, many languages lack an equivalent to the English word \emph{cousin}, and instead employ several (up to~16) more specific words that distinguish male and female cousins, elder and younger, paternal and maternal, etc. The ignorance of such semantic diversity in language processing can lead to hard-to-detect meaning-level mistakes. Today's top machine translation systems, for instance, make consistent mistakes over simple sentences such as \textit{My brother is younger than me} even across relatively high-resource languages: for example, the Japanese translation of the English sentence above is \begin{CJK}{UTF8}{min}私の兄は私より若いです\end{CJK} with a nonsensical meaning of \textit{My elder brother is younger than me.} The same mistake is observed over many other languages such as Mongolian or Hungarian.


It is clear from the example above that manifestations of language diversity are only revealed explicitly in a cross-lingual context, and remain hidden as long as one's point of view remains monolingual. Hence, the way to formalise lexical diversity is through building \emph{typology-informed multilingual lexical resources}. In this paper we exploit existing knowledge from lexical typology to enrich and extend existing multilingual lexicons which, in turn, can be reused in NLP applications. The key notion to capture lexical diversity is that of the \textit{lexical gap}, which refers to the lack of lexicalization of a particular concept in a particular language. 
We provide a formal approach to infer lexical gaps semi-automatically from existing domain-specific lexical knowledge. We apply the method to the domain of \emph{kinship}, well known for its cross-lingual diversity, building a resource that provides 37,370 gaps in 699 languages. Finally, as an example application in the context of multilingual NLP, we demonstrate how our resource can be used to evaluate machine translation systems over semantically challenging corpora. To our knowledge, ours is the first systematic method and the first lexical resource that provides lexical gaps for multiple languages in a large-scale and exhaustive manner.



The rest of the paper is organized as follows. We start by defining the notion of lexical gap and giving a brief review of lexical typology in Section~\ref{sec:our_proposal}. Our typology-based lexical gap generation method is described in Section~\ref{sec:method}. Section \ref{sec:evaluation} evaluates generation results based on native speaker input. Sections~\ref{sec:result} and~\ref{sec:application} present the resulting dataset and an example application for improving machine translation. Finally, we review related work and provide conclusions in sections~\ref{sec:soa} and~\ref{sec:conclusion}.

%% file: sections/3.lexgap.tex


The notion of \emph{lexical gap} is closely related to that of \textit{untranslatability} \cite{catford1978linguistic}. The latter, however, is a practically-oriented concept with many possible definitions: the acceptability of a translation depends on how rigorous one is regarding precision, conciseness, register, style, etc. Accordingly, the notion of lexical gap can be defined in broader or narrower ways. Linguists tend to adopt a more strict definition where monomorphemic words for which an equally monomorphemic translation cannot be provided are considered as gaps. For example, \newcite{wierzbicka2008there} considered that the concept of ``color'' is a lexical gap in Warlpiri, an Australian Indigenous language, as it lacks a word for it. Another example from \cite{ian2021etymology}, the general concept of ``rice'' is a lexical gap in Korean, which instead has words for more specific concepts ``cooked rice'' and ``uncooked rice''. 
In a computational context, \newcite{bentivogli2000looking} adopt a somewhat more relaxed definition, distinguishing translations into single words and \emph{restricted collocations} on the one hand, and \emph{free combinations of words} on the other hand, considering only the former as valid lexicalizations. Under this account, a restricted collocation is a stable combination of words (e.g., ``elder brother''). On the other hand, a free combination of words is a combination of words that are not bound together, and its parts can be freely used with other lexical items (e.g., ``father's elder brother''). In our research, we adopted the definition of \newcite{bentivogli2000looking} and, accordingly, considered a concept to be a lexical gap in a language if it can only be expressed through a free combination of words. Another criterion was only to consider as lexicalizations general-language words understood by ``average'' native speakers, as opposed to specialized terminology or rare and unknown words. For example, the gender-independent English \emph{nibling} and French \emph{adelphe} are specialized terms that designate \emph{nephew or niece} and \emph{sibling}, respectively. As these neologisms are only understood by specialists and are never used by the general public, we considered them as lexical gaps.

Lexical typology, a field of linguistics, explores systematic variations in the presence or absence of lexicalizations in languages with respect to specific domains (also known as ``domain categorization'') \cite{koch2001lexical}. English and Northern Sami, for example, categorize uncle-like relationships differently: Northern Sami has three different words with three distinct meanings---\textit{eahki} ``father's elder brother,'' \textit{čeahci} ``father's younger brother,'' and \textit{eanu} ``mother's brother''---while English packs all these meanings into the single term \textit{uncle}. 

Studies in lexical typology have mostly been conducted on domains that offer an unexpected richness of cross-lingual diversity: body parts, color, kinship, perception verbs, motion events, spatial dimension terms, cardinal direction terms, cutting and breaking events, putting and taking events, or pain predicates (for a review, see \newcite{koptjevskaja2015semantics} and \newcite{arora2021bhasacitra}).
However, only a small part of related scientific results have been published as actual datasets: \newcite{murdock1970kin}'s kinship categorization is published in D-PLACE \citelanguageresource{kirby2016d}. Parts of \newcite{brown1976general} and \newcite{kay2009world}'s works on colors are published under the lexicon chapter of WALS. To the best of our knowledge, no other works have been published as open resources.

More recently, digital lexical resources and parallel corpora have been increasingly used in lexical typology, enabling typologists to cover more languages and domains. For example, \newcite{viberg2014verbs}'s 50-language study on the perception domain was extended to 1,220 languages in \newcite{georgakopoulos2021universal}. A study of the color domain in 119~languages by \newcite{kay2009world} was leveraged in \newcite{mccarthy2019modeling}, which managed to cover 2,491~languages. Recent automated attempts to obtain lexico-semantic knowledge from large-scale parallel corpora provided promising results as well \cite{gast2018areal,levshina2021corpus}. Such bottom-up, data-driven approaches extend existing knowledge on lexical typology which, in turn, broaden the applicability of our method.

%% file: sections/4.method.tex

\begin{figure*}[ht]
\begin{center}
\includegraphics[width=\textwidth]{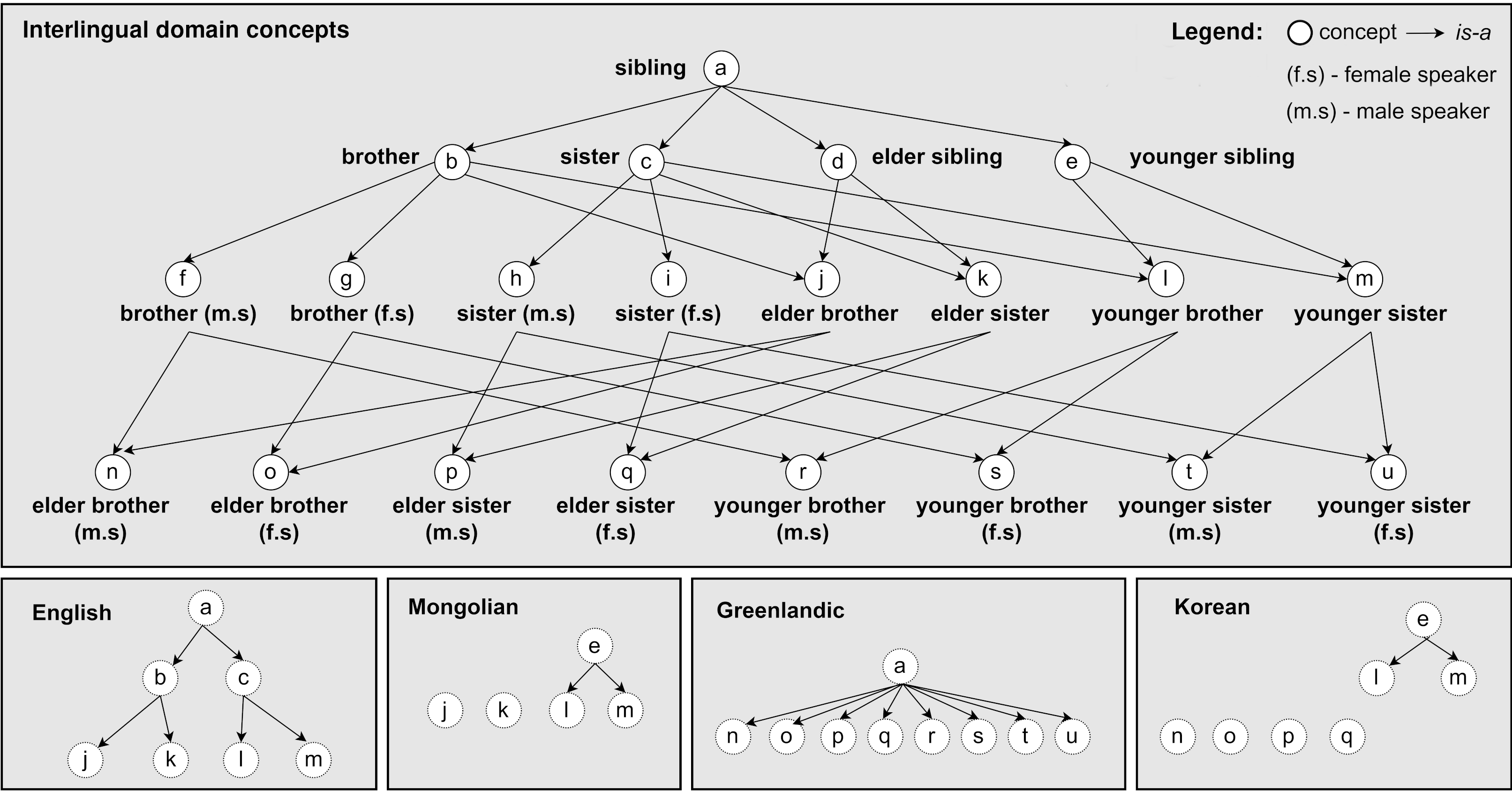} 
\caption{Interlingual conceptual layer of sibling domain.}
\label{fig:concept_core}
\end{center}
\end{figure*}

In this section, we describe a top-down, linguistically informed method for the systematic generation of lexical gaps for a large number of languages. While the method is generic and can be applied to multiple domains, our efforts so far have concentrated on the domain of kinship relations, from which all of our examples are taken. The method consists of the following three steps:
\begin{enumerate}
    \item \emph{domain specification:} the coverage of the study is defined in terms of domains and subdomains;
    \item \emph{conceptual modeling:} the domain selected is formalised through an interlingual hierarchy of lexical concepts;
    \item \emph{gap generation:} for each language covered by typological data, non-lexicalized concepts are marked as lexical gaps. 
\end{enumerate}

\subsection{Domain specification}

This step defines the domain of interest, based on the availability of typological data (lexicalization patterns) for the languages to be covered. The larger the domain and its subdomains, the more complex it is in the subsequent step to provide an interlingual representation. We chose to represent the \textit{kinship} domain as it is a crucial part of the general lexicon, yet it is known to be enormously diverse across languages and cultures. More precisely, we covered the six subdomains of \textit{grandparents}, \textit{grandchildren}, \textit{siblings}, \textit{uncles and aunts}, 
\textit{nephews and nieces}, 
and \textit{cousins}. We relied on the seminal work of \newcite{murdock1970kin} for lexicalization patterns and for a general characterization of the domain.

\subsection{Conceptual Modeling}

Modeling the interlingual conceptual space is an essential part of inferring lexical gaps in a systematic way. A purely conceptual approach to modeling the interlingua---developing or adopting an actual domain ontology---while feasible with limited effort and linguistic knowledge, produces models that are overly complex with sparse mappings to languages (e.g.~providing lots of non-lexicalized concepts). A purely linguistically motivated approach, instead, would produce results close to actual language use but would be extremely onerous, as it is based on the observation of potentially thousands of languages.

We propose a hybrid linguistic--conceptual approach where a top-down domain ontology is constrained by bottom-up linguistic data provided by the typological literature. In practice, for the kinship domain, we relied on \newcite{murdock1970kin} as an authoritative source of linguistic domain knowledge. This work provided the key formal attributes to consider for building the interlingua: \emph{gender of the kin} and \emph{of the speaker} (male, female, undefined), \emph{age of the kin} relative to the sibling and to the speaker (older, younger, undefined), \emph{relationship} (parent, child, or sibling). In two subdomains out of six (nephew/niece and cousin), information from typology literature was incomplete: for these subdomains, we relied on complementary information sources, namely native speaker input and Wiktionary words.

From these attributes, we automatically generated an exhaustive set of concepts and \emph{is-a} relations, such as \emph{parent's male sibling} or \emph{male parent's male sibling's female child (i.e.~father's brother's daughter) as pronounced by a female speaker}. The first concept is lexicalised in English as \emph{uncle} while the second one, a specific type of \emph{cousin} relationship, in Aleut as \emph{agiitudaxtanax}. There are, however, many theoretical combinations that are not actually lexicalized in any language. In order not to generate concepts that unnecessarily increase the size and complexity of the interlingua, we filtered the set of concepts generated to include only those attested by typological literature and native speaker input. As a result, we identified 198~concepts and 347~\textit{is-a} relations covering the six subdomains. A simplified example of the coverage of the \emph{sibling} subdomain can be seen in the top layer of Figure~\ref{fig:concept_core}, with the bottom layer showing existing lexicalizations in four languages.

\begin{table}[h]
    \centering
    \newcolumntype{R}{>{\raggedleft\arraybackslash}X}
\newcolumntype{C}{>{\centering\arraybackslash}X}
\newcolumntype{L}{>{\raggedright\arraybackslash}X}
    \begin{tabularx}{\linewidth}{lrRr}
    \hline
        Source data     & Languages & Words & Subdomains  \\\hline
        \newcite{murdock1970kin}  & 566       & 0  & 4           \\
        Wiktionary      & 166       & 1,681  & 6           \\
        Native speakers & 10        & 230      & 6     \\\hline
    \end{tabularx}
    \caption{Types of data used to infer lexical gaps.}
    \label{tab:types_data}
\end{table}

\subsection{Gap Generation}

For each concept defined above, this last step verifies whether it is lexicalized in each language covered, and if not, a lexical gap in that language is generated. In the kinship domain, we used three sources of evidence for lexicalization: (1)~Murdock's lexicalization patterns, (2)~Wiktionary words, and (3)~direct input from native speakers. Table \ref{tab:types_data} summarizes the types of data used to infer lexical gaps. (1)~provided lexicalization patterns for 566~languages, which we retrieved from \newcite{kemp2012kinship}. To give an example of a lexicalization pattern, the Javanese language belongs to the \emph{Algonkian sibling pattern}, which means that it lexicalizes the sibling terms: \emph{elder brother}, \emph{elder sister}, and \emph{younger sibling}. Although this database does not provide the actual words, it can be used to infer gaps. (2)~From Wiktionary, we extracted kinship terms in 166~languages, 144 of which were complementary to the Murdock dataset, using a custom-written parsing script. From the terms extracted, we automatically inferred the presence of gaps using the inference rules presented below. (3)~Lastly, for the languages Arabic, English, French, Hindi, Hungarian, Italian, Kannada, Malayalam, Mongolian, and Spanish, we asked two native speakers per language to provide lexicalization and gap information for all concepts. This information was used for evaluation (see Section~4 below) but also as a gold-standard source of terms and gaps for languages where incomplete or no data was available.


As Wiktionary does not explicitly indicate lexical gaps, we used the following rules to infer the presence of gaps from existing lexicalizations. The rules were derived from empirical patterns observed in the typology literature. 
\begin{description}
    \item[For concepts with speaker gender and age undefined:] if neither a concept~$c$ nor its parents have a lexicalization in language~$l$ then $c$ is a lexical gap in~$l$.
    \item[For concepts with speaker gender or age specified:] if language~$l$ is known not to indicate the speaker's gender or age in the lexicalization, then all concepts with these attribute are lexical gaps in~$l$.
\end{description}

We explain the first rule through the example of \emph{uncle}. If a language lexicalizes \emph{uncle}, it might also express the more specific \emph{paternal} and \emph{maternal uncle}, e.g.,~by adding appropriate adjectives. We cannot automatically infer that \emph{paternal uncle} and \emph{maternal uncle} are gaps: 
deciding whether collocations are restricted is far from trivial, as even native speakers may disagree on the everyday usage of expressions such as \emph{paternal uncle}, \textit{younger sibling}, or \textit{female cousin}.\footnote{The use of corpus-based frequency data is promising future work in this direction.} We consider, however, that complex expressions corresponding to indirect hyponyms (of distance~2 or more), such as \emph{father's elder brother} or \emph{mother's younger sister}, are never restricted collocations and can therefore be considered as gaps.

The second rule is explained by the rarity of the phenomenon of incorporating the speaker's gender and age into lexicalizations (e.g.~a male Ewe person would call his grandchild differently than a female Ewe would). It is safe to assume that languages that do not possess these properties have no concise way of expressing this information as part of the kinship terminology.

%% file: sections/5.evaluation.tex
Our evaluations verified the correctness and completeness of the automated word extraction and gap generation method described in Section~\ref{sec:method}, and also extended the overall coverage of our resource. Our goal was to obtain highly reliable and reusable data on lexical diversity, we thus gave priority to precision over recall. Furthermore, as we considered our sources from linguistic typology as \emph{a priori} reliable, we instead focussed our evaluations on the words and gaps that were automatically inferred from Wiktionary data. Thus, we verified the correctness of: (a)~the original Wiktionary data, generally considered as reliable, nevertheless not fully error-free; (b)~our Wiktionary data extraction logic; (c)~our lexical gap inference rules.

\subsection*{Evaluation Setup}

The evaluation was based on input provided by language speakers in ten languages: Arabic, English, French, Hindi, Hungarian, Italian, Kannada, Malayalam, Mongolian, and Spanish. Two speakers per language were employed: the first one a native speaker born and educated (university-level) within the speaker community, while the second speaker was every time a language expert with (at least) proficiency in the language and a good knowledge of lexical semantics. The first speaker provided initial input which was subsequently verified and extensively discussed with the expert speaker in order to ensure the coherence of the input and to avoid misunderstandings. We also made sure that native speakers received clear prior instructions on what is meant by a concept being lexicalized or not in a language. These instructions covered the following principles:
\begin{itemize}
    \item \emph{general vs specialized language:} we only consider terms that sound natural in general spoken language: thus, \emph{sibling} is accepted but \emph{nibling}, extremely rare and only known to specialists, is not;
    \item \emph{restricted collocations vs free combinations:} fixed expressions that are frequently used are acceptable as lexicalizations (e.g.~\emph{little brother}), while expressions that are not felt as fixed and frequent should be considered as gaps (e.g.~\emph{female cousin});
    \item \emph{usage context:} speakers were encouraged to signal (in writing) stylistic, dialectal, or other constraints of usage for the words provided.  
\end{itemize}

Native speakers were provided with the full concept list of every subdomain, with concepts described in English (all contributors were fluent English speakers), such as \emph{``elder sister's child (as pronounced by a female speaker)''}. They were also given lexicalizations extracted from Wiktionary wherever available, as well as indications of lexical gaps that we automatically inferred. They were asked to validate these words and gaps, and also to provide lexicalizations and flag gaps for any concept not covered by Wiktionary. In the case of an incorrect word, they either had to provide a correct word or had to indicate it as a gap. In the case of an incorrect gap, they had to provide the appropriate word.

Language-specific words extracted from Wiktionary for eight out of these ten languages were sufficient to infer gaps. In Malayalam and Kannada, Wiktionary provided insufficient input and thus could not be used for gap inference. Ultimately, 165~words were retrieved by Wiktionary, 1,059~gaps were automatically inferred, and 230~words and 444~gaps were provided by native speakers as in Table~\ref{tab:eval_setup}.

\begin{table}[h]
\centering
\newcolumntype{R}{>{\raggedleft\arraybackslash}X}
\newcolumntype{C}{>{\centering\arraybackslash}X}
\newcolumntype{L}{>{\raggedright\arraybackslash}X}
\begin{tabularx}{\linewidth}{lrrrr}
\hline
Language & Wiktionary & Inferred & Expert & Expert  \\
         &  words     & gaps       &  words     & gaps \\\hline
Arabic    & 22        & 126        & 6            & 36           \\
English   & 16        & 134        & 16           & 20           \\
French    & 20        & 129        & 16           & 29           \\
Hindi     & 38        & 124        & 7            & 16           \\
Hungarian & 22        & 127        & 13           & 28           \\
Italian   & 16        & 136        & 10           & 24           \\
Kannada   & 1         & 0          & 63           & 128          \\
Malayalam & 3         & 0          & 60           & 131          \\
Mongolian & 12        & 144        & 23           & 7            \\
Spanish   & 16        & 139        & 16           & 25           \\
\hline
Total     & 165       & 1,059      & 230          & 444          \\
\hline
\end{tabularx}
\caption{Statistics of the evaluation data.}
\label{tab:eval_setup}
\end{table}


\subsection*{Evaluation results}
Our gap inference rules were conservative by design in order to favor precision, which is reflected in our evaluation results. Precision over inferred lexical gaps is very high, in the~99--100\% range both across languages (Table~\ref{tab:evaluation1}) and subdomains (Table~\ref{tab:evaluation2}). Gap recall is~85.1\%. False positive gaps only occurred in Hungarian and Mongolian. Hungarian \textit{nagyszülő} ``grandparent'' was absent from Wiktionary, and was subsequently assumed as a lexical gap. As only one term  {\fontencoding{T2A}\selectfont үеэл} `cousin' was retrieved for Mongolian, our first gap inference rule assumed that all indirect descendants (of distance~2 or higher) were lexical gaps. Mongolian speakers informed us, however, that {\fontencoding{T2A}\selectfont үеэл ах} ``elder male cousin'' and {\fontencoding{T2A}\selectfont үеэл эгч} ``elder female cousin'' were widely used collocations.

\begin{table}[h]
\centering
\newcolumntype{R}{>{\raggedleft\arraybackslash}X}
\newcolumntype{C}{>{\centering\arraybackslash}X}
\newcolumntype{L}{>{\raggedright\arraybackslash}X}
\begin{tabularx}{\linewidth}{lrRr}
\hline
Domain                        & Languages & Gaps & Cohen's Kappa                 \\
\hline
grandparents            & 19        & 246  & 0.94                   \\
grandchildren           & 8         & 135  & 0.98                   \\
siblings                & 2         & 20   & 0.90                   \\
uncle/aunt                & 17        & 335  & 0.75                   \\
\hline
Total                   & 22        & 772  & 0.89        \\\hline          
\end{tabularx}
\caption{For each subdomain, the agreement between, on the one hand, our Wiktionary-inferred and native-verified data and, on the other hand, typological evidence from Murdock. The second and third columns provide the data size used for evaluation, in terms of the number of overlapping languages and gaps.}
\label{tab:kappa}
\end{table}

The overlap between gaps signalled by Murdock and inferred by our method consisted of 772~gaps in 22~languages (i.e.~merely~2\% of the entire gap dataset, see Table \ref{tab:kappa}). We computed the agreement between the two data sources and, using Cohen's Kappa, we obtained a score of~0.89, further implying that the gaps we inferred are high quality.



\begin{table}[h]
\centering
\small
\newcolumntype{R}{>{\raggedleft\arraybackslash}X}
\newcolumntype{C}{>{\centering\arraybackslash}X}
\newcolumntype{L}{>{\raggedright\arraybackslash}X}
\begin{tabularx}{\linewidth}{l|rRR|rRR}
\hline
\multirow{2}{*}{Languages} & \multicolumn{3}{c|}{Words} & \multicolumn{3}{c}{Gaps} \\
                            & \multicolumn{1}{c}{\textbf{$P$}}       & \multicolumn{1}{c}{\textbf{$R$}}      & \multicolumn{1}{c|}{\textbf{$F_1$}}      & \multicolumn{1}{c}{\textbf{$P$}}       & \multicolumn{1}{c}{\textbf{$R$}}      & \multicolumn{1}{c}{\textbf{$F_1$}}     \\
\hline
Arabic                     & 100.0   & 78.6   & 88.0   & 100.0   & 77.8   & 87.5  \\
English                    & 100.0   & 50.0   & 66.7   & 100.0   & 87.0   & 93.0  \\
French                     & 80.0    & 50.0   & 61.5   & 100.0   & 81.6   & 89.9  \\
Hindi                      & 100.0   & 84.4   & 91.5   & 100.0   & 88.6   & 94.0  \\
Hungarian                  & 100.0   & 62.9   & 77.2   & 99.2    & 81.8   & 89.7  \\
Italian                    & 100.0   & 61.5   & 76.2   & 100.0   & 85.0   & 91.9  \\
Mongolian                  & 100.0   & 34.3   & 51.1   & 98.6    & 95.3   & 96.9  \\
Spanish                    & 93.8    & 48.4   & 63.9   & 100     & 84.8   & 91.8  \\
\hline
Total                      & 96.9    & 59.5   & 73.7   & 99.7    & 85.1   & 91.8 \\
\hline
\end{tabularx}

\caption{Native speaker evaluation of words and lexical gaps by language.}
\label{tab:evaluation1}
\end{table}

\begin{table}[h]
\centering
\small
\newcolumntype{R}{>{\raggedleft\arraybackslash}X}
\newcolumntype{C}{>{\centering\arraybackslash}X}
\newcolumntype{L}{>{\raggedright\arraybackslash}X}
\begin{tabularx}{\linewidth}{L|rrr|rrr}
\hline
\multirow{2}{*}{Subdomains} & \multicolumn{3}{c|}{Words} & \multicolumn{3}{c}{Gaps} \\
                            & \multicolumn{1}{c}{\textbf{$P$}}       & \multicolumn{1}{c}{\textbf{$R$}}      & \multicolumn{1}{c|}{\textbf{$F_1$}}      & \multicolumn{1}{c}{\textbf{$P$}}       & \multicolumn{1}{c}{\textbf{$R$}}      & \multicolumn{1}{c}{\textbf{$F_1$}}     \\
\hline
grandparents                & 93.3    & 66.7   & 77.8   & 99.0    & 80.5   & 88.8  \\
grandchildren               & 100.0   & 71.4   & 83.3   & 100.0   & 85.3   & 92.1  \\
siblings                    & 91.7    & 56.9   & 70.2   & 100.0   & 88.3   & 93.8  \\
uncle/aunt                    & 100.0   & 47.3   & 64.2   & 100.0   & 83.2   & 90.8  \\
nephew/niece                   & 100.0   & 63.6   & 77.8   & 100.0   & 85.3   & 92.1  \\
cousins                     & 100.0   & 60.4   & 75.3   & 99.4    & 86.1   & 92.3  \\
\hline
Total                       & 96.9    & 59.5   & 73.7   & 99.7    & 85.1   & 91.8 \\
\hline
\end{tabularx}

\caption{Native speaker evaluation of words and lexical gaps by domain.}
\label{tab:evaluation2}
\end{table}

As for existing lexicalisations: the overall precision over Wiktionary words (after extraction) was~96.9\%. The slight loss of precision is due to the appearance of rare French and Spanish terms in Wiktionary. Thus, our French evaluators noted that \textit{adelphe} for ``sibling'' is a newly coined and mostly unknown term (e.g.~it does not appear in the 1996 edition of the \emph{Robert}). Moreover, \textit{aïeul} and \textit{aïeule} were identified as obsolete for designating ``grandfather''  and ``grandmother''. \textit{Cadet} was judged not to mean ``younger sibling'', as stated by Wiktionary, but rather ``younger brother.'' Likewise, the Spanish gender-neutral \textit{hermane} ``sibling'' is a very rarely used neologism according to our evaluators. Apart from these examples, the terms retrieved from Wiktionary turned out to be of very high quality.

Recall on words is relatively low at 59.5\%, which is a sign of Wiktionary incompleteness. Native speakers (other than Malayalam and Kannada) provided 107~new words and collocations (Table~\ref{tab:eval_setup}). In particular, many missing terms are expressed through restricted collocations in English, Spanish, Mongolian, and Hungarian. For example, 23~new inputs made by Mongolian speakers were all restricted collocations; e.g.~{\fontencoding{T2A}\selectfont ач хүү} ``son's son,'' { \fontencoding{T2A}\selectfont нагац ах} ``maternal uncle.'' Other examples are English \textit{elder sister}, Spanish \textit{hermano menor} ``younger brother'' or \textit{tío materno} ``maternal uncle,'' Hungarian \textit{fiúunoka} ``grandson'' or \textit{nagytestvér} ``elder sibling.'' In addition, our French speakers identified some words or morphological alternations missing from Wiktionary, e.g.~the colloquial French \textit{tata} ``aunt,'' \textit{papi} ``grandfather,'' but also \textit{aîné} ``elder brother''.


\begin{table*}[ht]
\setlength{\tabcolsep}{5pt}
\centering
\begin{tabular}{l|rr|rr|rrr|rr}
\hline
\multirow{2}{*}{Domain} & \multirow{2}{*}{Concepts} & \multirow{2}{*}{Relations}  & \multicolumn{2}{c|}{Murdock} & \multicolumn{3}{c|}{Wiktionary + Expert} & \multicolumn{2}{c}{Total}  \\
                        &                           &        & Languages & Gaps            & Languages & Words & Gaps               & Languages & Gaps           \\\hline
grandparents            & 19                        & 31                                           & 459       & 6,137           & 99    & 391      & 1,280             & 539       & 7,171          \\
grandchildren           & 27                        & 55                                        & 183       & 3,763           & 72    & 202         & 1,457              & 247       & 5,049          \\
siblings                & 21                        & 33                                        & 162       & 1,762           & 145    & 498        & 2,109              & 304       & 3,851          \\
uncles/aunts                & 31                        & 51                                          & 559       & 15,188          & 83    & 312       & 1,650              & 625       & 16,503         \\
nephews/nieces                & 33                        & 47                                          & n/a       & n/a             & 65     & 214      & 1,606               & 65        & 1,606           \\
cousins                 & 67                        & 130                                        & n/a       & n/a             & 60    & 294        & 3,190              & 60        & 3,190          \\\hline
Total                   & 198                       & 347                                      & 561       & 26,850          & 168    & 1,911       & 11,292             & 699       & 37,370         \\\hline
\end{tabular}
\caption{Statistics of the lexical gap resource.}
\label{tab:results}
\end{table*}
\bigbreak

\begin{table*}[t]
\setlength{\tabcolsep}{3pt}
\begin{tabular}{lp{5.5cm}p{9.5cm}}
\hline
File & Descrption & Columns \\\hline
Concepts & lexical kinship concepts in interlingua & \emph{subdomain}, \emph{concept label}, \emph{description}, \emph{provenance} \\
Relations & hypernymy relations across concepts & \emph{subdomain}, \emph{hypernym concept label}, \emph{hyponym concept label} \\
Words & lexicalizations in supported languages & \emph{subdomain}, \emph{concept label}, \emph{lang name}, \emph{ISO code}, \emph{term}, \emph{provenance} \\
Gaps & lexical gaps in supported languages & \emph{subdomain}, \emph{concept label}, \emph{lang name}, \emph{ISO code}, \emph{evidence} \\\hline
\end{tabular}
\caption{\label{tab:structure}Tab-separated files composing the kinship resource and their attributes.}
\end{table*}

%% file: sections/6.result.tex
Based on evaluation results, we considered the precision of our gap inference rules to be high enough to be applied to the final dataset without any modification. They were re-run over the lexicalizations that were manually corrected based on native speaker input.

Statistics on the final resource obtained are provided in Table~\ref{tab:results}. The resource, as in Figure~\ref{fig:concept_core}, is organized into a lexico-semantic interlingua layer and a layer of language-specific lexicons. The interlingua represents the six kinship subdomains through a total of 198~concepts and 347~hypernymy relations.
In the lexicon layer, we automatically inferred 37,370~lexical gaps in 699~languages. 1,911~words were retrieved from Wiktionary and native speaker inputs. We inferred 26,850~gaps in 561~languages from Murdock data and 11,292~gaps in 168~languages from Wiktionary and native speaker inputs.

The resource is described online\footnote{\url{http://ukc.disi.unitn.it/index.php/kinship/}} and is also directly available for download.\footnote{\url{ https://github.com/kbatsuren/KinDiv}} It is distributed as four tab-separated text files, the structure of which is described in Table~\ref{tab:structure}.
The \emph{concept label} column holds formal, structured representations of kinship concepts, such as \texttt{x;El;Br;Ch} meaning \emph{elder brother's child as pronounced by a male speaker} (the last attribute is indicated by \texttt{x}). The \emph{provenance} and \emph{evidence} columns, in turn, provide the origin of the information, which can be: a reference to the typology research data, a reference to a lexical data source (e.g.~Wiktionary), or \emph{``native speaker.''}

\begin{figure}
    \centering
    \includegraphics[width=\columnwidth]{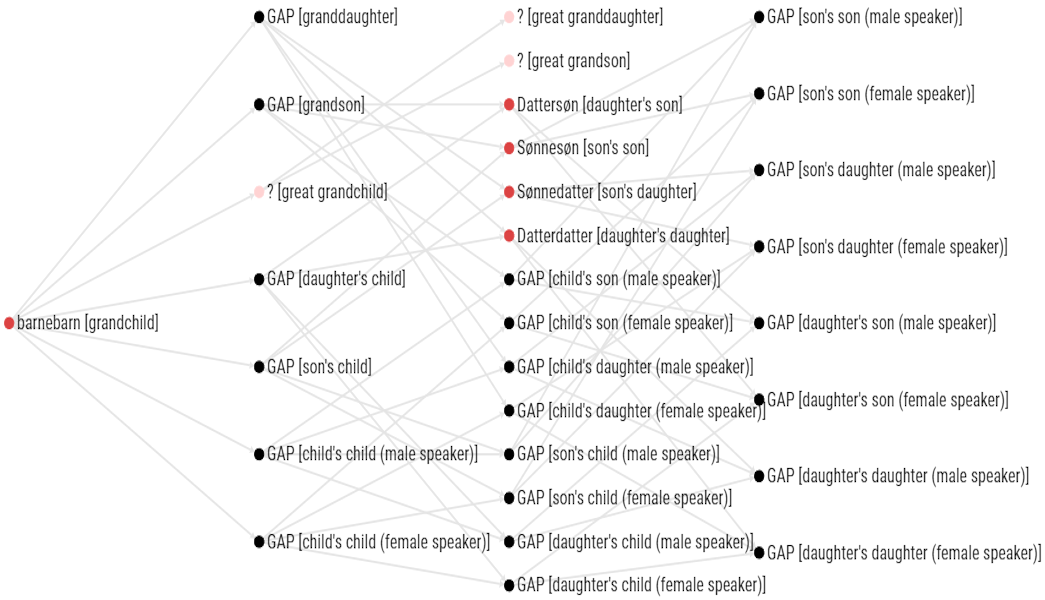}
    \caption{Interactive browser tool, showing lexicalizations and gaps for the grandchildren subdomain in Danish.}
    \label{fig:ukc_kinship}
\end{figure}

The kinship database has also been imported into the \emph{Universal Knowledge Core} (UKC)\footnote{\url{http://ukc.datascientia.eu}}, a multilingual lexical database of more than 1,000~languages \cite{giunchiglia2017understanding,giunchiglia2018one}. The focus of the UKC is language diversity: it is capable of representing lexical gaps and its online version is equipped with interactive visualisation tools that allow the browsing of kinship terms and gaps by subdomain in all supported languages \cite{bella2022language}, as shown in Figure~\ref{fig:ukc_kinship}.



%% file: sections/7.application.tex
We believe that resources on lexical gaps have a high potential in improving existing cross-lingual applications, for example as pivots in multilingual representation learning, as seeds in cross-lingual model transfer, and also as part of multilingual word embeddings. In this section, we demonstrate how our resource can be used to improve and evaluate machine translation systems. 

\subsection*{The Gap Problem in Machine Translation}

Lexico-semantic diversity makes both human and machine translation challenging. Due to the lack of a concise translation equivalent for a given word or expression, translators need to find a term without excessively corrupting the original meaning. The following cases from the kinship domain demonstrate the challenge.


The sentences below are real machine translation (MT) outputs for the English sentence \emph{My brother is three years younger than me} in five languages:\footnote{Obtained from Google Translate on 20 December 2021.}

\vspace{5pt}
{
\setlength{\tabcolsep}{5pt}
\begin{tabular}{ll}
    \hline
    Target lang & Target sentence \\\hline
    Hungarian & *A bátyám három évvel fiatalabb nálam. \\
    Japanese  & *\begin{CJK}{UTF8}{min}私の兄は私より3歳年下ですす.\end{CJK}\\
    Korean    & *\begin{CJK}{UTF8}{mj}형은 나보다 세 살 아래다.\end{CJK}\\
    Mongolian & {\fontencoding{T2A}\selectfont *Ах маань надаас гурван насаар дүү.}\\
    Russian   & {\fontencoding{T2A}\selectfontМой брат младше меня на три года.}\\\hline
\end{tabular}
}
\vspace{5pt}

The only correct output is in Russian, due to the English \textit{brother} having the equivalent {\fontencoding{T2A}\selectfont брат}. On the other hand, \textit{brother} is a lexical gap in Hungarian, Japanese, Korean, and Mongolian, as these languages all express the relationship with more specific words. A corpus-statistics-based translation approach leads to a severe meaning-level mistake, as the Hungarian \textit{bátyám}, the Japanese \begin{CJK}{UTF8}{min}兄\end{CJK}, and the Mongolian {\fontencoding{T2A}\selectfont ах} all mean `elder brother,' leading to the nonsensical result of \emph{My elder brother is three years younger than me.} The Korean case is even worse as, besides assuming an elder brother, it also implies that the speaker is necessarily male. 

In case of a non-existent equivalent, a semantically informed translation method can choose a broader term (e.g. \emph{sibling}) instead of a narrower one, achieving approximation at the expense of slight information loss, less critical than inadvertently injecting false information. This works for Hungarian (\emph{testvér}) but not for Mongolian that has no equivalent for \emph{sibling} either, and where an even broader term such as \emph{relative} may not sound right. In such cases, the appropriate narrower term (between {\fontencoding{T2A}\selectfont ах} ``elder brother'' or {\fontencoding{T2A}\selectfont эрэгтэй дүү} ``younger brother'') could, in this particular case, be inferred from the sentence context by a human translator or a sophisticated automated method. 

The explicit and formal representation of untranslatability, as offered by our kinship resource, can be exploited to improve translation systems, but also to evaluate the semantic performance of existing systems on challenging tasks. In the rest of the section we illustrate the evaluation scenario through an experimental case study. 

\subsection*{Machine Translation Evaluation Method}

We describe a \emph{semantic} evaluation measure and method for MT systems, designed to capture meaning-level translation mistakes that conventional metrics (e.g.~the BLEU score) are known not to address adequately \cite{wu2016google}. The method consists of:
\begin{enumerate}
    \item building a semantically annotated benchmark corpus of hard-to-translate sentences;
    \item translating the sentences into a pre-defined set of target languages using an automated MT system;
    \item measuring the \emph{semantic distance} between key terms in the original and translated sentences.
\end{enumerate}
A formal lexical gap resource, such as the one presented in this paper, is used in step~1, for the construction and annotation of sentences, and also in step~3 for the computation of semantic distances, based on the \emph{least common subsumer distance} between the original and the translated term meaning in the interlingual concept hierarchy.

As a case study on step~1, we built a benchmark corpus of 50~English sentences that contain kinship terms, by adapting sentences from the British National Corpus \citelanguageresource{burnard1995users}. The sentences contain 7--9 representative terms from each kinship subdomain: each such term appearing in a sentence was annotated by its meaning (in terms of the corresponding interlingual concept). Due to the well-known pervasiveness of lexical diversity among kinship terms, we consider this corpus as an effective meaning-level evaluation set of hard-to-translate sentences.

In step~2, we translated the 50~sentences into five languages: Hungarian, Japanese, Korean, Mongolian, and Russian, using Google Translate.

In step~3, we measured the semantic distance between target words and translation outputs. First, we automatically disambiguated output words (e.g.~the Mongolian {\fontencoding{T2A}\selectfont ах} was disambiguated as ``elder brother'', formally represented as \texttt{El;Br}). Then we computed the semantic distance between this and the gold standard annotation. Figure~\ref{fig:concept_core} shows that the Mongolian {\fontencoding{T2A}\selectfont ах} and {\fontencoding{T2A}\selectfont эрэгтэй дүү} are connected through the concept of ``brother.'' From ``elder brother'' to ``younger brother'' the distance is two hops, so the semantic distance amounts to~2. 

\subsection*{Machine Translation Experiment Results}

Table \ref{tab:mt_eval} shows for each language pair the number of gaps, the average semantic distance over gap-containing sentences, and the average semantic distance over all sentences. Machine translation of kinship terms turned out to be much more robust from English into Russian than into Japanese, Korean, Hungarian, or Mongolian, and a likely explanation for that is the fewer gaps encountered in Russian than the others. 

On the whole, one can also observe that the gap-based distances are higher than the overall distances, which proves that handling gaps is indeed a weak point of current MT techniques that is likely worth addressing via dedicated solutions.


\begin{table}[h]
\setlength{\tabcolsep}{3pt}
\centering
\newcolumntype{R}{>{\raggedleft\arraybackslash}X}
\newcolumntype{C}{>{\centering\arraybackslash}X}
\newcolumntype{L}{>{\raggedright\arraybackslash}X}
\begin{tabularx}{\linewidth}{lrrr}
\hline
Language pair & Gaps & Sem.dist (gaps) & Sem.dist (all)  \\\hline
English--Russian   & 6    & 1.00                & 0.34                \\
English--Japanese  & 13   & 1.06                   & 0.38                \\
English--Korean    & 12   & 1.58                & 0.90                 \\
English--Hungarian & 19   & 1.31                & 1.06                \\
English--Mongolian & 12   & 1.33                & 1.12                \\
\hline
\end{tabularx}
\caption{Semantic evaluation of Google Translate from English towards five languages.}
\label{tab:mt_eval}
\end{table}


%% file: sections/2.related.tex
The large-scale multilingual lexical databases that exist today have been, by and large, produced and used by two distinct communities of researchers, namely historical and computational linguists. The former community has produced the \emph{Intercontinental Dictionary Series} (IDS) \citelanguageresource{ids}, the \emph{Automated Similarity Judgement Program} (ASJP) \citelanguageresource{wichmann2013asjp}, \emph{CLICS} \citelanguageresource{list2017database}, \emph{CLDF} \cite{forkel2018cross}, and the \emph{World Loanword Database} (WOLD) \citelanguageresource{wold}. These resources typically consist of phonemic descriptions of words or transliterations, as modern orthographies are irrelevant to both comparative and historical linguistics. This characteristic, however, makes these resources difficult to use in computational applications that target contemporary written language.

The computational linguist and NLP communities, on the other hand, rely on resources derived from and describing contemporary written language. Formal, computer-processable lexical databases, such as BabelNet \citelanguageresource{navigli2012babelnet}, the Open Multilingual Wordnet (OMW) \citelanguageresource{bond2013linking}, or Concepticon \citelanguageresource{list2016concepticon}, however, focus on representing sameness, i.e.~word meanings shared across languages, and do not explicitly indicate untranslatability. In BabelNet and OMW, language-specific word meanings either are left out or are mapped to other languages in an approximative manner. Such inaccuracies lead to a Western-centric bias as the word meanings that are correctly mapped belong to dominant languages such as the English Princeton WordNet \citelanguageresource{miller1995wordnet}.

We are aware of two efforts that address the formal representation and methodology of building diversity-aware lexical databases: MultiWordNet (MWN) \citelanguageresource{pianta2002multiwordnet} and the announced second version of the Open Multilingual Wordnet (OMW2) \citelanguageresource{bond2020some}. MWN has inbuilt support for representing lexical gaps and provides about~300 and 1,000 lexical gaps in Hebrew \citelanguageresource{ordan2007hebrew} and Italian \cite{bentivogli2000looking}, respectively. The recent paper \citelanguageresource{bond2020some} announced that OMW2 would be based on the \emph{Collaborative InterLingual Index} (CILI), which envisions a collaborative method for defining language-specific concepts and gaps. We are not yet aware of the availability of an actual resource that would correspond to the theoretical diversity-aware representational abilities of the CILI.
As also put forth in \newcite{bentivogli2000looking} and \newcite{ordan2007hebrew}, identifying lexical gaps in a systematic manner is far from trivial. Let us take the example of the English word \emph{cousin} which has no equivalent in Hindi and the Hindi word \raisebox{-1ex}{\includegraphics[height=0.5cm]{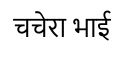}} meaning \emph{son of father's brother} which, in turn, has no concise equivalent in English. The Hindi gap corresponding to \emph{cousin} can be identified as part of a expert-driven lexicon translation effort---called \emph{expansion} in \newcite{fellbaum2012challenges}---that is frequently used to produce lexicons for relatively lower-resourced languages. This approach was used, for example, to provide around 600~gaps in the Unified Scottish Gaelic Wordnet \citelanguageresource{bella2020major} and 79~gaps in Mongolian WordNet \citelanguageresource{ganbold2014experiment,batsuren2019building}. This method, however, is effort-intensive and does not provide any gaps in the reverse direction. Traditional bilingual dictionaries that explicitly indicate untranslatability may be a good source of gaps, which corresponds to the \emph{merge} method used in MWN. Such high-quality dictionaries, however, are not available for lots of languages.

Our approach is different from both of these: instead of the entire lexicon, it focuses on a single domain that is well-known to be cross-lingually diverse, such as kinship, colors, or body parts. Instead of human experts or existing lexical resources, it relies on evidence from linguistic typology. As a result, in the given domain it provides a much more exhaustive coverage both in terms of the number of gaps per language and in terms of the number of languages covered. Lastly but perhaps most importantly, ours is a predominantly top-down method that, thanks to knowledge from linguistic typology, is based on a prior conceptual understanding of the domain at hand (e.g.~for the kinship domain, the analysis of 162~languages by \newcite{murdock1970kin}). The result is that we are able to construct a hierarchy of interlingual lexico-semantic domain concepts with considerable precision, with language-specific lexicalisations (or the lack thereof) easily mappable to the hierarchy for lots of languages. In contrast, the bottom-up \emph{expand} and \emph{merge} methods, envisaged for the CILI and used in MWN, proceed by gradually extending an English-specific hierarchy, as new languages and new concepts are ``encountered''. This leads to the need for a (sometimes non-monotonic) reorganisation of the concept graph as the knowledge about a given domain evolves based on cross-lingual evidence. As shown in Figure~\ref{fig:concept_core}, the one-by-one addition of Mongolian, Greenlandic, and Korean leads to profound changes in the interlingual representation.

%% file: sections/8.conclusion.tex
Our paper and the corresponding resource aim to address a gap in current multilingual lexicons and cross-lingual applications, namely the representation and exploitation of lexical diversity. We formally capture diversity through the notions of language-specific concepts and lexical gaps, and provide a systematic method to produce such data in a semi-automated manner. Our first large-scale effort focused on the domain of kinship terminology, well known to be particularly diverse across languages and cultures. The resulting machine-readable resource provides a wide coverage of domain concepts and languages (198~kinship concepts covered in 699~languages), and is freely available both for online browsing and download. In the future, we plan to apply the method presented in this paper to formalize new domains that are known to be diverse, such as colors, food, or visual objects \cite{giunchiglia2021classifying}. One such ongoing project concerns culturally specific concepts in the languages of India \cite{nair2022indoukc}. Finally, we believe that resources such as ours provide essential information to lexically-focused cross-lingual applications, such as multilingual language models or cross-lingual transfer. We have presented one such application in the context of machine translation, but we plan to explore other application areas in the future. We also plan to extend our machine translation experiments to additional state-of-the-art MT systems.